\documentclass[twocolumn,10,a4paper]{article}
\newcommand{\gbf}[1] {\mbox{\boldmath${#1}$\unboldmath}}
\newcommand{\be}{\begin{equation}}
\newcommand{\ee}{\end{equation}}
\newcommand{\beq}{\begin{equation}}
\newcommand{\eeq}{\end{equation}}
\newcommand{\bed}{\begin{displaymath}}
\newcommand{\eed}{\end{displaymath}}
\newcommand{\beqa}{\begin{eqnarray}}
\newcommand{\eeqa}{\end{eqnarray}}
\newcommand{\beqann}{\begin{eqnarray*}}
\newcommand{\eeqann}{\end{eqnarray*}}
\newcommand{\bseq}{\begin{subequations}}
\newcommand{\eseq}{\end{subequations}}

\newcommand{\ba}{\begin{array}}
\newcommand{\ea}{\end{array}}

\newcommand{\negr}[1]{{\bf {#1}}}

\usepackage[dvips]{graphicx}
\usepackage[dvips]{floatflt}
\usepackage{subeqn}

\begin{document}
\title{
\vskip -10mm
\parbox{174mm}{
\begin{flushright}
{\small Proceedings of the $11$th World Congress in Mechanism and
Machine Science \\ August $18$-$21$, $2003$, Tianjin, China \\ China
Machinery Press, edited by Tian Huang \\}
\end{flushright}} 
\vskip 9pt {\bf \Large A Comparative Study between
\goodbreak Two Three-DOF Parallel Kinematic Machines using Kinetostatic Criteria and
 Interval Analysis}\vskip 12pt
{\normalsize Damien Chablat$^1$  Philippe Wenger$^1$ Jean-Pierre Merlet $^2$ \\
 \normalsize $^1$Institut de Recherche en Communications et Cybern\'etique de Nantes \\
 \normalsize 1, rue de la No\"e, 44321 Nantes, France, \\
 \normalsize $^2$INRIA Sophia-Antipolis,
 \normalsize 2004 Route des Lucioles, \\
 \normalsize 06902 Sophia Antipolis, France, \\
 \normalsize Damien.Chablat@irccyn.ec-nantes.fr and Jean-Pierre.Merlet@sophia.inria.fr\\}
\vskip -27pt}
\date{}
\maketitle
\pagestyle{plain}\thispagestyle{empty}
\hskip -3.5mm {\small \bf Abstract:} {\small This paper addresses
the workspace analysis of two 3-DOF translational parallel
mechanisms designed for machining applications. The two machines
features three fixed linear joints. The joint axes of the first
machine are orthogonal whereas these of the second are parallel.
In both cases, the mobile platform moves in the Cartesian $x-y-z$
space with fixed orientation.  The workspace analysis is conducted
on the basis of prescribed kinetostatic performances. Interval
analysis based methods are used to compute the dextrous workspace
and the largest cube enclosed in this workspace.}

\section*{{\normalsize ${\bf 1}$ Introduction}}
\vskip -11pt
Parallel kinematic machines (PKM) are known for their high dynamic
performances and low positioning errors. The kinematic design of
PKM has drawn the interest of several researchers. The workspace
is usually considered as a relevant design criterion
\cite{Merlet:96,Clavel:88,Gosselin:91}. Parallel singularities
\cite{Wenger:97ICAR} occur in the workspace where stiffness is
lost, and thus are very undesirable. They are generally eliminated
by design. The Jacobian matrix, which relates the joint rates to
the output velocities is generally not constant and not isotropic.
Consequently, the performances (e.g. maximum speeds, forces,
accuracy and stiffness) vary considerably for different points in
the Cartesian workspace and for different directions at one given
point. This is a serious drawback for machining applications
\cite{Kim:1997,Treib:1998,Wenger:1999b}. Some parallel mechanisms
were recently shown to be isotropic throughout the workspace
\cite{Kong:2002,Carricato:2002,Kim:2002}. But their legs are
subject to bending which is not desirable for machining
applications . To be of interest for machining applications, a PKM
should preserve good workspace properties, that is, regular shape
and acceptable kinetostatic performances throughout. In milling
applications, the machining conditions must remain constant along
the whole tool path \cite{Rehsteiner:1999a,Rehsteiner:1999b}. In
many research papers, this criterion is not taken into account in
the algorithmic methods used to calculate the workspace volume
\cite{Luh:1996,Merlet:1999}.
\par
This paper compares two PKM with three translational DOF derived
from the Delta robot originally designed by Reymond Clavel for
pick-and-place operations \cite{Clavel:88}. The first one, called
UraneSX (Renault Automation) \cite{Company:2000}, has three non
coplanar horizontal linear joints, like the Quickstep (Krause \&
Mauser) . The second one, called Orthoglide, has three orthogonal
linear joints \cite{Chablat:2002}.
\par
The comparative study is conducted on the basis of the size of a
prescribed workspace with bounded velocity and force transmission
factors, called the dextrous workspace. Interval analysis based
method is used to compute the dextrous workspace as well as the
largest cube enclosed in this workspace \cite{Merlet:2000}.
\par
Next section presents the Orthoglide and UraneSX mechanisms. The
kinematic equations and the singularity analysis are detailed in
Section~3. Section~4 is devoted to the determination of the
largest cube enclosed in the dextrous workspace and to the
comparative study between the two mechanisms.
\vspace*{-13pt}
\section*{{\normalsize ${\bf 2}$ Description of the Orthoglide and the UraneSX}}
\vspace*{-5pt}
Most existing PKM can be classified into two main families. PKM of
the first family have fixed foot points and variable length struts
and are generally called ``hexapods''. PKM of the second family
have variable foot points and fixed length struts. They are
interesting because the actuators are fixed and thus the moving
masses are lower than in the hexapods and tripods.

\begin{figure}[!hu]
    \begin{center}
           \centerline{\hbox{\includegraphics[width=65mm]{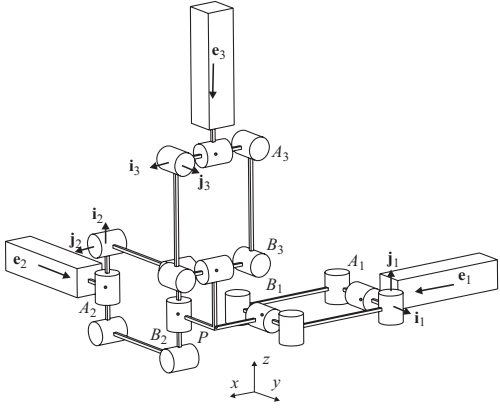}}}
           \caption{{\bf Orthoglide kinematic architecture}}
           \protect\label{figure:Orthoglide}
    \end{center}
\end{figure}

\begin{figure}[!hb]
    \begin{center}
           \centerline{\hbox{\includegraphics[width=45mm]{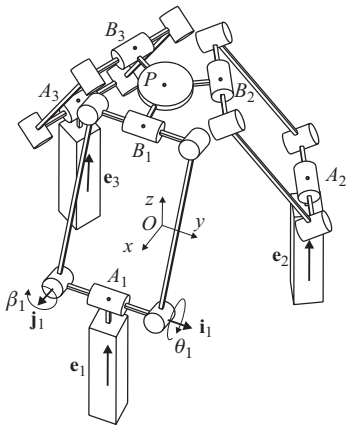}}}
           \caption{{\bf UraneSX kinematic architecture}}
           \protect\label{figure:UraneSX}
    \end{center}
\end{figure}

The Orthoglide and the UraneSX mechanisms studied in this paper
are $3$-axis translational PKM and belong to the second family.
Figures~\ref{figure:Orthoglide} and \ref{figure:UraneSX} show the
general kinematic architecture of the Orthoglide and of the
UraneSX, respectively. Both mechanisms have three parallel $PRPaR$
identical chains (where $P$, $R$ and $Pa$ stand for Prismatic,
Revolute and Parallelogram joint, respectively). The actuated
joints are the three linear joints.

The output body is connected to the linear joints through a set of
three parallelograms of equal lengths $L~=~A_iB_i$, so that it can
move only in translation. Vectors $\negr e_i$ coincide with the
direction of the $i$th linear joint. The base points $A_i$ are
located at the middle of the first two revolute joints of the
$i^{th}$ parallelogram, and $B_i$ is at the middle of the last two
revolute joints of the $i^{th}$ parallelogram.

For the Orthoglide mechanism, the first linear joint axis is
parallel to the $x$-axis, the second one is parallel to the
$y$-axis and the third one is parallel to the $z$-axis. When each
vector $\negr e_i$ is aligned with $\negr A_i \negr B_i$, the
Orthoglide is in an isotropic configuration and the tool center
point $P$ is located at the intersection of the three linear joint
axes.

The linear joint axes of the UraneSX mechanism are parallel to the
$z$-axis. In fig.~\ref{figure:UraneSX}, points $A_1$, $A_2$ and
$A_3$ are the vertices of an equilateral triangle whose geometric
center is $O$ and such that $OA_i=R$. Thus, points $B_1$, $B_2$
and $B_3$ are the vertices of an equilateral triangle whose
geometric center is $P$, and such that $OB_i=r$.
\section*{{\normalsize ${\bf 3}$ Kinematic Equations and Singularity Analysis}}
\vspace*{-5pt}
We recall briefly here the kinematics and the singularities of the
Orthoglide and of the UraneSX (See
\cite{Company:2000,Chablat:2002} for more details).
\subsection*{{\normalsize ${\bf 3.1}$ Kinematic equation and singularity analysis}}
\vskip -5pt
Let $\theta_i$ and $\beta_i$ denote the joint angles of the
parallelogram about  axes $\negr i_i$ and $\negr j_i$,
respectively (Figs.~\ref{figure:Orthoglide} and
\ref{figure:UraneSX}). Let $\negr \rho_1$, $\negr \rho_2$, $\negr
\rho_3$ denote the linear joint variables and $L$ denote the
length of the three legs, $A_iB_i$.

For the Orthoglide, the position vector \negr p of the tool center
point $P$ is defined in a reference frame (O, $x$, $y$, $z$)
centered at the intersection of the three linear joint axes (note
that the reference frame has been translated in
Fig.~\ref{figure:Orthoglide} for more legibility).

For the UraneSX, the position vector \negr p of the tool center
point $P$ is defined in a reference frame (O, $x$, $y$, $z$)
centered at the geometric center of the points $A_1$, $A_2$, and
$A_3$ (same remark as above).

Let $\dot{\gbf {\rho}}$ be referred to as the vector of actuated
joint rates and $\dot{\negr p}$ as the velocity vector of point
$P$:
 \bed
    \dot{\gbf{\rho}}=
    [\dot{\rho}_1~\dot{\rho}_2~\dot{\rho}_3]^T
   ,\quad
    \dot{\negr p}=
    [\dot{x}~\dot{y}~\dot{z}]^T
 \eed
$\dot{\negr p}$ can be written in three different ways by
traversing the three chains $A_iB_iP$:
 \be
    \dot{\negr p} =
    \negr e_i \dot{\rho}_i +
    (\dot{\theta}_i \negr i_i + \dot{\beta}_i \negr j_i)
    \times
    (\negr b_i - \negr a_i) \\
   \label{equation:cinematique}
 \ee
where $\negr b_i$ and $\negr c_i$ are the position vectors of the
points $B_i$ and $C_i$, respectively, and $\negr e_i$ is the
direction vector of the linear joints, for $i=1, 2, $3.

We want to eliminate the three idle joint rates $\dot{\theta}_i$
and $\dot{\beta}_i$ from Eqs.~(\ref{equation:cinematique}), which
we do by dot-multiplying Eqs.~(\ref{equation:cinematique}) by
$\negr b_i - \negr a_i$:
 \be
   (\negr b_i - \negr a_i)^T \dot{\negr p} =
   (\negr b_i - \negr a_i)^T
   \negr e_i \dot{\rho}_i
 \label{equation:cinematique-2}
 \ee
Equations (\ref{equation:cinematique-2}) can now be cast in vector
form, namely $\negr A \dot{\bf p} = \negr B \dot{\gbf \rho}$,
where \negr A and \negr B are the parallel and serial Jacobian
matrices, respectively:
 \beqa
   \negr A =
   \left[\begin{array}{c}
           (\negr b_1 - \negr a_1)^T \\
           (\negr b_2 - \negr a_2)^T \\
           (\negr b_3 - \negr a_3)^T
         \end{array}
   \right]
   {\rm ~~and~~}
   \negr B =
   \left[\begin{array}{ccc}
            \eta_1&
            0 &
            0 \\
            0 &
            \eta_2&
            0 \\
            0 &
            0 &
            \eta_3
         \end{array}
   \right]
 \label{equation:A_et_B}
 \eeqa
with $\eta_i= (\negr b_i - \negr a_i)^T \negr e_i $ for $i=1,2,3$.
\par
Parallel singularities occur when the determinant of the matrix
\negr A vanishes, {\it i.e.} when $\det(\negr A)=0$.
Eq.~(\ref{equation:A_et_B}) shows that the parallel singularities
occur when:
 \bed
    (\negr b_1 - \negr a_1) =
    \alpha  (\negr b_2 - \negr a_2) +
    \lambda (\negr b_3 - \negr a_3)
 \eed
that is when the points $A_1$, $B_1$, $A_2$, $B_2$, $A_3$ and
$B_3$ lie in parallel planes. A particular case occurs when the
links $A_iB_i$ are parallel:
 \beqa
    (\negr b_1 - \negr a_1) //
    (\negr b_2 - \negr a_2)
    \quad {\rm and} \quad
    (\negr b_2 - \negr a_2) //
    (\negr b_3 - \negr a_3) \\
    \quad {\rm and} \quad
    (\negr b_3 - \negr a_3) //
    (\negr b_1 - \negr a_1)\nonumber
 \eeqa
Serial singularities arise when the serial Jacobian matrix \negr B
is no longer invertible {\it i.e.} when $\det(\negr B)=0$. At a
serial singularity a direction exists along which no Cartesian
velocity can be produced. Equation~(\ref{equation:A_et_B}) shows
that $\det(\negr B)=0$ when for one leg $i$, $(\negr b_i - \negr
a_i) \perp (\negr b_i - \negr a_i)$.

When \negr A is not singular, we can write,
  \be
   \dot{\negr p} = \negr J  \dot{\gbf \rho} {\rm ~with~}
   \negr J = \negr A^{-1} \negr B
  \ee
\subsection*{{\normalsize ${\bf 3.2}$ Velocity transmission factors}}
\vskip -5pt
For joint rates belonging to a unit ball, namely, $||\dot{\gbf
\rho}|| \leq 1$, the Cartesian velocities belong to an ellipsoid
such that:
 \bed
   \dot{\gbf p}^T (\negr J \negr J^T) \dot{\gbf p} \leq 1
 \eed
The eigenvectors of matrix $\negr J \negr J^T$ define the
direction of the principal axes of this ellipsoid. The square
roots $\psi_1$, $\psi_2$ and $\psi_3$ of the real eigenvalues
$\sigma_1$, $\sigma_2$ and $\sigma_3$ of $\negr J \negr J^T$, {\it
i.e.} the lengths of the aforementioned principal axes are the
velocity transmission factors in the principal axes directions. To
limit the variations of these factors in the Cartesian workspace,
we set
 \beqa
   \psi_{min} \leq \psi_i \leq \psi_{max}
   \label{e:velocity_limits}
 \eeqa
throughout the workspace. To simplify the problem, we set
$\psi_{min}=1/ \psi_{max}$ where the value of $\psi_{max}$ depends
on given performance requirements.
\section*{{\normalsize ${\bf 4}$ Determination of the largest cube enclosed in the dextrous
workspace}}
\vspace*{-5pt}
For usual machine tools, the Cartesian workspace shape is
generally a parallelepiped. Due to the symmetrical architecture of
the Orthoglide, the Cartesian workspace has a fairly regular shape
in which it is possible to include a cube whose sides are parallel
to the planes $xy$, $yz$ and $xz$ respectively.

The Cartesian workspace of the UraneSX is the intersection of
three cylinders whose axes are parallel to the $z$-axis. Thus, the
workspace is theoretically unlimited in the $z$-direction and the
Jacobian matrix does not depend on the $z$ coordinate.
Practically, only the limits on the linear joints define the
limits of the Cartesian workspace in the $z$-directions.

The aim of the following section is to define the edge length of
the largest cube enclosed in the dextrous workspace of the
Orthoglide and the edge length of the largest square enclosed in
the dextrous workspace of the UraneSX, as well as the location of
their respective centers for both mechanisms. This is done using
an interval analysis method. Unlike numerical computing methods,
such a method allows to prove formally that the velocity
amplification factors lie in the predefined range
$[\psi_{min}~\psi_{max}]$ in a given subpart of the Cartesian
workspace.

Two algorithms are described to define (i) a set of boxes in the
Cartesian workspace in which the velocity amplification factors
remain under the prescribed values, and (ii) the largest cube
enclosed in the dextrous workspace.
\subsection*{{\normalsize ${\bf 4.1}$ Box verification}}
\vskip -5pt
A basic tool of the algorithm is a module ${\cal M}(B)$ that takes
as input a Cartesian box $B$ and whose output is:
\begin{itemize}
\item either that for any point in the box the eigenvalues lie in the
range $[\sigma_{min},\sigma_{max}]$
\item or that for any point in the box one of the eigenvalues is either
lower than $\sigma_{min}$ or larger than $\sigma_{max}$
\item or that the two previous conditions does not hold for all the
points of the box {\it i.e.} that for some points the eigenvalues
lie in the range $[\sigma_{min},\sigma_{max}]$ while this is not
true for some other points
\end{itemize}
The first step of this module consists in considering an arbitrary
point of the box (e.g. its center) and to compute the eigenvalues
at this point: either all of them lie in the range
$[\sigma_{min},\sigma_{max}]$ or at least one of them lie outside
this range.

In the first case if we are able to check that there is no point
in $B$ such that one of the eigenvalues at this point is be equal
to $\sigma_{min}$ or $\sigma_{max}$, then we can guarantee that
for any point in $B$ the eigenvalues will be in the range
$[\sigma_{min},\sigma_{max}]$. Indeed assume that at a given point
$B$ the lowest eigenvalue is lower than $\sigma_{min}$: this
implies that somewhere along the line joining this point to the
center of the box the lowest eigenvalue will be exactly
$\sigma_{min}$. To perform this check we substitute $\sigma_{min}$
to the unknown in the characteristic polynomial of $\negr J \negr
J^T$  to get a polynomial in $x, y, z$ only. We now have to check
if there exists some values for these three cartesian coordinates
that cancel the polynomial, being understood that these values
have to define a point belonging to $B$: this is done by using an
interval analysis algorithm from the {\tt ALIAS}
library~\cite{Merlet:2000}.

Assume now that at the center of the box the largest eigenvalue is
greater than $\sigma_{max}$. If there is no point in $B$ such that
one of the eigenvalues is equal to $\sigma_{max}$, then we can
guarantee that for any point in $B$ the largest eigenvalue will
always be greater than $\sigma_{max}$. This check is performed by
using the same method as in the previous case. Hence the ${\cal
M}$ module will return:
\begin{itemize}
\item 1: if for all points in $B$ the eigenvalues lie in
$[\sigma_{min},\sigma_{max}]$ (hence $B$ is in the dextrous
workspace)
\item -1: if for all points in $B$ either the largest eigenvalue is
always greater than $\sigma_{max}$ or the lowest eigenvalue is
lower than $\sigma_{min}$ (hence $B$ is outside the dextrous
workspace).
\item 0: in the other cases {\it i.e.} parts of $B$ may be either outside or
inside the dextrous workspace
\end{itemize}
\subsection*{{\normalsize ${\bf 4.2}$ Determination of the dextrous workspace}}
\vskip -5pt
The dextrous workspace ${\cal W}$ is here defined as the loci of
the points for which square real roots of the eigenvalues  of the
matrix $\negr J \negr J^T$, {\it i.e.} the velocity transmission
factors, lie within the predefined range
$[\sigma_{min},\sigma_{max}]$. The eigenvalues are determined by
solving the third degree characteristic polynomial of the matrix
$\negr J \negr J^T$.

The polynomial is only defined for the points within the
intersection ${\cal I}$ of the three cylinders defined by
 \bed
 x^2+y^2 \le L \quad
 x^2+z^2 \le L \quad
 y^2+z^2 \le L
 \eed
for the Orthoglide, and,
 \beqa
 (x-R+r)^2+y^2 &\le& L \nonumber \\
 \left(x-(R-r)/2\right)^2+\left(y-(R-r)\sqrt{3}/2\right)^2 &\le& L \nonumber \\
 \left(x-(R-r)/2\right)^2+\left(y+(R-r)\sqrt{3}/2\right)^2 &\le& L \nonumber
 \eeqa
for the UraneSX.
\par
To solve numerically the above equations and to compare the two
mechanisms, the length of the legs is normalized, {\it i.e.} we
set $L=1$.
\subsection*{{\normalsize ${\bf 4.3}$ Algorithm}}
\vskip -5pt
We will now describe a method for determining  a cube that is
enclosed in the workspace of both mechanisms, whose edge length is
$2W$ and such that there is no other cube enclosed in the
workspace with an edge length of $2(W+\alpha)$, where $\alpha$ is
an accuracy threshold fixed in advance.

The first step is to determine the largest cube enclosed in the
workspace with a center located at (0,0,0). This is done by using
the ${\cal M}$ module on the Cartesian box $B_{init}$
$[-k\alpha,k\alpha],[-k\alpha,k\alpha], [-k\alpha,k\alpha]$ where
$k$ is an integer initialized to 1. Each time the ${\cal M}$
module returns 1 for $B_{init}$ (which means that the cube with
edge length $2k\alpha$ is enclosed in the dextrous workspace) we
double the value of $k$. If this module returns -1 for a value of
$k$ larger than 1 this implies that the cube with edge length
$k\alpha/2$ lie in the dextrous workspace while the cube with edge
length $k\alpha$ does not. Hence we restart the process with
$k=k/2+1$.Otherwise we have determined that the cube with edge
length $2k\alpha$ is enclosed in the dextrous workspace, while the
cube with edge length $2(k+1)\alpha$ does not. The value
$2k\alpha$ is hence an initial value for $W$.

We then use the following algorithm for determining the largest
cube enclosed in the dextrous workspace. We start with the
Cartesian box $B_0$= $[-L,L]$, $[-L,L]$, $[-L,L]$ that enclose the
workspace and we use a list of Cartesian boxes ${\cal L}$ indexed
by $i$ with $n$ elements:
\begin{enumerate}
\item If $i > n$ EXIT
\item Using interval arithmetics, check if  $B_i$
contains points such that a box centered at these points with edge
length $W+\alpha$ is fully enclosed in the workspace.
\begin{enumerate}
\item  If the box does not contain such points discard the box,
set $i$ to $i+1$ and restart.
\item If the box contains points that belong and point that does not
belong to the workspace, check if the box has at least one range
that is larger than $\alpha$. If not, there is no point in the box
that can be the center of a cube with edge length at least
$W+\alpha$ and that is enclosed in the workspace. Hence we may
discard this box. If yes bissect the box along one of this range,
thereby creating 2 new boxes that are stored at the end of ${\cal
L}$. Set $i$ to $i+1$ and restart.
\end{enumerate}
\item At this stage the box $B_i$ contains only points that may be the
center of a cube with edge length at least $W+\alpha$ and fully
enclosed in the workspace.
\item If the maximal width of the box is lower or equal to $\alpha$, use
the procedure described for the center at (0,0,0) to determine the
largest cube centered at the center  of the box that lie within
the dextrous workspace. If this procedure provides a cube with
edge length larger than $W$, update $W$. Set $i$ to $i+1$ and
restart.
\item If the maximal width of the box is greater than $\alpha$ use the
procedure described for the center at (0,0,0) to determine the
largest cube centered at the center  of the box that lie within
the dextrous workspace. If this procedure provides a cube with
edge length larger than $W$, update $W$. Select a variable of the
box that has a range larger than $\alpha$, bissect the box,
thereby creating 2 new boxes that are stored at the end of ${\cal
L}$. Set $i$ to $i+1$ and restart.
\end{enumerate}
This procedure ensures to determine a cube with edge length $W$
that is enclosed in the workspace and in the dextrous workspace,
while there is no such cube with edge length $W+\alpha$.
\subsection*{{\normalsize ${\bf 4.4}$ Design parameters and results}}
\vskip -5pt
To compare the two mechanisms, the leg length $L$ is set to 1 and
the bounds on the velocity factor amplification are $\psi=[0.5~2]$
with $\alpha=0.001$. For the UraneSX, it is necessary to define
two additional lengths, $r$ and $R$. However, the edge length of
the workspace only depends on $R-r$.

For the Orthoglide, we found out that the largest cube has its
center located at $(0.086, 0.086, 0.086)$ and that its edge length
is $L_{Workspace}= 0.644$.

For the UraneSX, the design parameters are those defined in
\cite{Company:2000}, which we have normalized to have $L=1$, {\it
i.e.} $r=3/26$ and $R= 7/13$. To compare the two mechanisms, we
increase the value of $R$ such that $R'=R+\lambda$ with
$\lambda=[0.0,0.2]$. For $R< 7/13$, the constraints on the
velocity amplification factors  are not satisfied.

\begin{table}
  \begin{center}
   \begin{tabular}{|c|c|c|} \hline
$\lambda$ & Center & $L_{Workspace}$ \\ \hline
 0.00 &   (-0.0178,-0.0045)&   0.510\\ \hline
 0.05 &   (-0.0179,-0.0022)&   0.470\\ \hline
 0.10 &   (-0.0225,-0.0031)&   0.420\\ \hline
 0.15 &   (-0.0245,-0.0018)&   0.370\\ \hline
 0.20 &   (-0.0211,-0.0033)&   0.320\\ \hline
   \end{tabular}
   \caption{{\bf Variations of the edge length of the square workspace for the UraneSX mechanism}}
  \end{center}
\end {table}

The optimal value of $R'$ is obtained for $\lambda=0$, {\it i.e.}
for the design parameters defined in \cite{Company:2000} for an
industrial application. To expand this square workspace in the
$z$-direction, the range limits must be equal to the edge length
of the square plus the range variations necessary to move
throughout the square in the $x-y$ plane.

The constraints on the velocity amplification factors used for the
design of the Orthoglide are close to those used for the design of
the UraneSX, which is an industrial machine tool. For the same
length of the legs, the size of the cubic workspace is larger for
the Orthoglide than for the UraneSX.

For the Orthoglide, the optimization puts the serial and parallel
singularities far away from the Cartesian workspace
\cite{Chablat:2002}. The UraneSX has no parallel singularities due
to the design parameters ($R-r < L$), but serial singularity
cannot be avoided with the previous optimization function. To
produce the motion in the $z$-direction, the range limits of the
linear joints are set such that the constraints on the velocity
amplification factors are not satisfied throughout the Cartesian
workspace.
\section*{{\normalsize ${\bf 5}$ Conclusions}}
\vspace*{-5pt}
Two 3-DOf translational PKM are compared in this paper: the
Orthoglide and the UraneSX. The dextrous workspace is defined as
the part of the Cartesian workspace where the velocity
amplification factors remain within a predefined range. The
dextrous workspace is really available for milling tool paths
because the performances are homogeneous in it. The largest cube
for the Orthoglide and the largest square for the UraneSX enclosed
in the dextrous workspace are computed using an interval analysis
based method. This method is interesting because it allows to
prove formally whether in a subpart of the Cartesian workspace the
velocity amplification factors remain within a predefined range or
not. These results can be used to design partially these
mechanisms for milling applications.
\section*{{\normalsize Acknowledgments }}
\vspace*{-5pt}
This research was supported by  CNRS, project ROBEA ``Machine
d'Architecture compleXe''.
 \vspace*{-6pt}
\def\refname{\large References}
\bibliographystyle{unsrt}

\begin{thebibliography}{99}
 \vspace*{-1pt}
\bibitem{Merlet:96}
Merlet, J-P., 1996,
\newblock ``Workspace-0riented Methodology for Designing a
Parallel Manipulator,''
\newblock IEEE Int. Conf. on Robotics and Automation, pp.~3726-3731.
 \vspace*{-6pt}
\bibitem{Clavel:88}
Clavel, R., 1988,
\newblock ``DELTA, a Fast Robot with Parallel Geometry,''
\newblock Proc. of the 18th Int. Symposium of Industrial Robots,
IFR Publications, pp. 91-100.
 \vspace*{-6pt}
\bibitem{Gosselin:91}
Gosselin, C and Angeles, J., 1991,
\newblock ``A Global Performance Index for the Kinematic
Optimization of Robotic Manipulators,''
\newblock Journal of Mechanical Design, vol.~113, pp.~220-226.
 \vspace*{-6pt}
\bibitem{Wenger:97ICAR}
Wenger, Ph. and Chablat, D., 1997,
\newblock ``Definition Sets for the Direct Kinematics of Parallel
Manipulators,''
\newblock 8th Int. Conf. Advanced Robotics, pp.~859-864.
 \vspace*{-6pt}
\bibitem{Kim:1997}
Kim, J.\, Park, C.\, Kim, J.\ and Park, F.C.\, 1997,
\newblock ``Performance Analysis of Parallel Manipulator Architectures for CNC Machining
Applications,''
\newblock Proc. IMECE Symp. On Machine Tools, Dallas.
 \vspace*{-6pt}
\bibitem{Treib:1998}
Treib, T.\ and Zirn, O.\,
\newblock ``Similarity laws of serial and parallel
manipulators for machine tools,''
\newblock Proc. Int. Seminar on Improving Machine Tool Performance, pp.~125--131, Vol.~1, 1998.
 \vspace*{-6pt}
\bibitem{Wenger:1999b} 
Wenger, P.\, Gosselin, C.\ and Chablat. D.\, 2001,
\newblock ``A Comparative Study
of Parallel Kinematic Architectures for Machining Applications,''
\newblock Proc. Workshop on Computational Kinematics'2001, Seoul, Korea, pp.~249--258.
 \vspace*{-6pt}
\bibitem{Kong:2002}
Kong, X. and Gosselin, C. M., 2002, ``A Class of 3-DOF
Translational Parallel Manipulators with Linear I-O Equations,''
Proc. of Workshop on Fundamental Issues and Future Research
Directions for Parallel Mechanisms and Manipulators, Québec,
Canada.
 \vspace*{-6pt}
\bibitem{Carricato:2002}
Carricato, M. and Parenti-Castelli, V., 2002, ``Singularity-Free
Fully-Isotropic Translational Parallel Mechanisms,'' The
International Journal of Robotics Research, Vol. 21, No. 2, pp.
161-174, February.
 \vspace*{-6pt}
\bibitem{Kim:2002}
Kim, H.S. and Tsai, L.W., 2002, ``Evaluation of a Cartesian
manipulator,'' in Lenar\v{c}i\v{c}, J. and Thomas, F. (editors),
{\em Advances in Robot Kinematic}, Kluwer Academic Publishers,
June, pp. ~21--38.
 \vspace*{-6pt}
\bibitem{Rehsteiner:1999a} 
Rehsteiner, F., Neugebauer, R.. Spiewak, S. and Wieland, F., 1999,
\newblock ``Putting Parallel Kinematics Machines (PKM) to Productive Work,''
\newblock Annals of the CIRP, Vol.~48:1, pp.~345--350.
 \vspace*{-6pt}
\bibitem{Rehsteiner:1999b} 
Tlusty, J., Ziegert, J, and Ridgeway, S., 1999,
\newblock ``Fundamental Comparison of the Use of Serial and Parallel Kinematics for Machine Tools,''
\newblock Annals of the CIRP, Vol.~48:1, pp.~351--356.
 \vspace*{-6pt}
\bibitem{Luh:1996} 
Luh  C-M., Adkins F. A., Haug E. J. and Qui C. C., 1996, ``Working
Capability Analysis of Stewart platforms,'' Trans. of ASME,
pp.~220--227.
 \vspace*{-6pt}
\bibitem{Merlet:1999} 
Merlet J-P., 1999, ``Determination of 6D Workspace of Gough-Type
Parallel Manipulator and Comparison between Different
Geometries,'' The Int. Journal of Robotic Research, Vol.~19,
No.~9, pp.~902--916.
 \vspace*{-6pt}
\bibitem{Company:2000} 
Company, O. Pierrot, F., 2002, ``Modelling and Design Issues of a
3-axis Parallel Machine-Tool,'' Mechanism and Machine Theory,
Vol.~37, pp.~1325--1345.
 \vspace*{-6pt}
\bibitem{Chablat:2002} 
Chablat, D. and Wenger, Ph, 2002, ``Architecture Optimization of a
3-DOF Parallel Mechanism for Machining Applications, the
Orthoglide,'' IEEE Trans. Robotics and Automation, June, 2003.
 \vspace*{-6pt}
\bibitem{Golub:1989}
Golub, G. H. and Van Loan, C. F., 1989, ``Matrix Computations,''
The John Hopkins University Press, Baltimore.
 \vspace*{-6pt}
\bibitem{Salisbury:1982}
Salisbury J-K. and Craig J-J., 1982, ``Articulated Hands: Force
Control and Kinematic Issues,'' The Int. J. Robotics Res., Vol.~1,
No.~1, pp.~4--17.
 \vspace*{-6pt}
\bibitem{Angeles:1997}
Angeles J., 2002, ``Fundamentals of Robotic Mechanical Systems,''
Second Edition, Springer-Verlag, New York.
\bibitem{Yoshikawa:1985}
Yoshikawa, T., 1985, ``Manipulability of Robot Mechanisms,'' The
Int. J. Robotics Res., Vol.~4, No.~2, pp.~3--9.
 \vspace*{-6pt}
\bibitem{Merlet:2000} 
Merlet J-P., 2000, ``{ALIAS}: an interval analysis based library
for solving and analyzing system of equations,'' SEA, June.
Automation, pp.~1964--1969.
\end{thebibliography}

\end{document}